\documentclass[runningheads]{llncs}
\usepackage[T1]{fontenc}
\usepackage{graphicx}
\usepackage{graphicx,verbatim}
\usepackage{amsmath}
\usepackage{amssymb}
\usepackage{float}
\usepackage{multirow}
\usepackage[table,xcdraw]{xcolor}
\usepackage{amssymb}
\usepackage{hyperref}

\newcommand{\methodname}{HyCLoST}
\newcommand{\uniname}{UNI2-h}

\begin{document}
\title{Hyperbolic Contrastive Learning with Entailment for Spatial Transcriptomics}
%
%
\author{Daniela Vega\thanks{Corresponding author: d.vegaa@uniandes.edu.co}\orcidID{0009-0002-9731-7591} \and
Paula Cárdenas\orcidID{0009-0005-1185-548X} \and
Hannah Ceballos\orcidID{0009-0005-5398-3770} \and
Leonardo Manrique\orcidID{0009-0008-9428-6009} \and
Pablo Arbeláez\orcidID{0000-0001-5244-2407}}


\authorrunning{D. Vega et al.}
\titlerunning{Hyperbolic Contrastive Learning for Spatial Transcriptomics}
%
\institute{Center for Research and Formation in Artificial Intelligence \\ Universidad de los Andes, Colombia \\
\email{\{d.vegaa,p.cardenasg,h.ceballos,dl.manrique,pa.arbelaez\}@uniandes.edu.co}}
\maketitle              
\begin{abstract}
Spatial Transcriptomics (ST) has transformed biomedical research by enabling the spatial mapping of gene expression across tissue sections. However, high operational costs, specialized equipment requirements, and sensitivity to experimental noise limit the accessibility and scalability of ST. Recent computer vision approaches aim to overcome these limitations by predicting spatial gene expression directly from histopathology images. While effective, current approaches often suffer from gene expression over-smoothing and overly uniform predictions across tissue regions, suggesting that further progress depends on learning representations that reflect the hierarchical and asymmetric structure of gene regulation and tissue morphology. To address these issues, we propose Hyperbolic Contrastive Learning with Entailment for Spatial Transcriptomics (\methodname{}), a hyperbolic contrastive learning model that captures the intrinsic hierarchical relationships within ST data. By leveraging hyperbolic geometry and a gene-to-image entailment loss, \methodname{} learns structured, biologically grounded representations that improve gene expression prediction accuracy, achieving a 6\% reduction in MSE and an 8\% increase in PCC across 26 ST datasets, over previous methods. Our source code is publicly available at \url{https://github.com/BCV-Uniandes/HyCLoST}.

\end{abstract}

\begin{figure*}[htbp]
\includegraphics[width=0.98\textwidth]{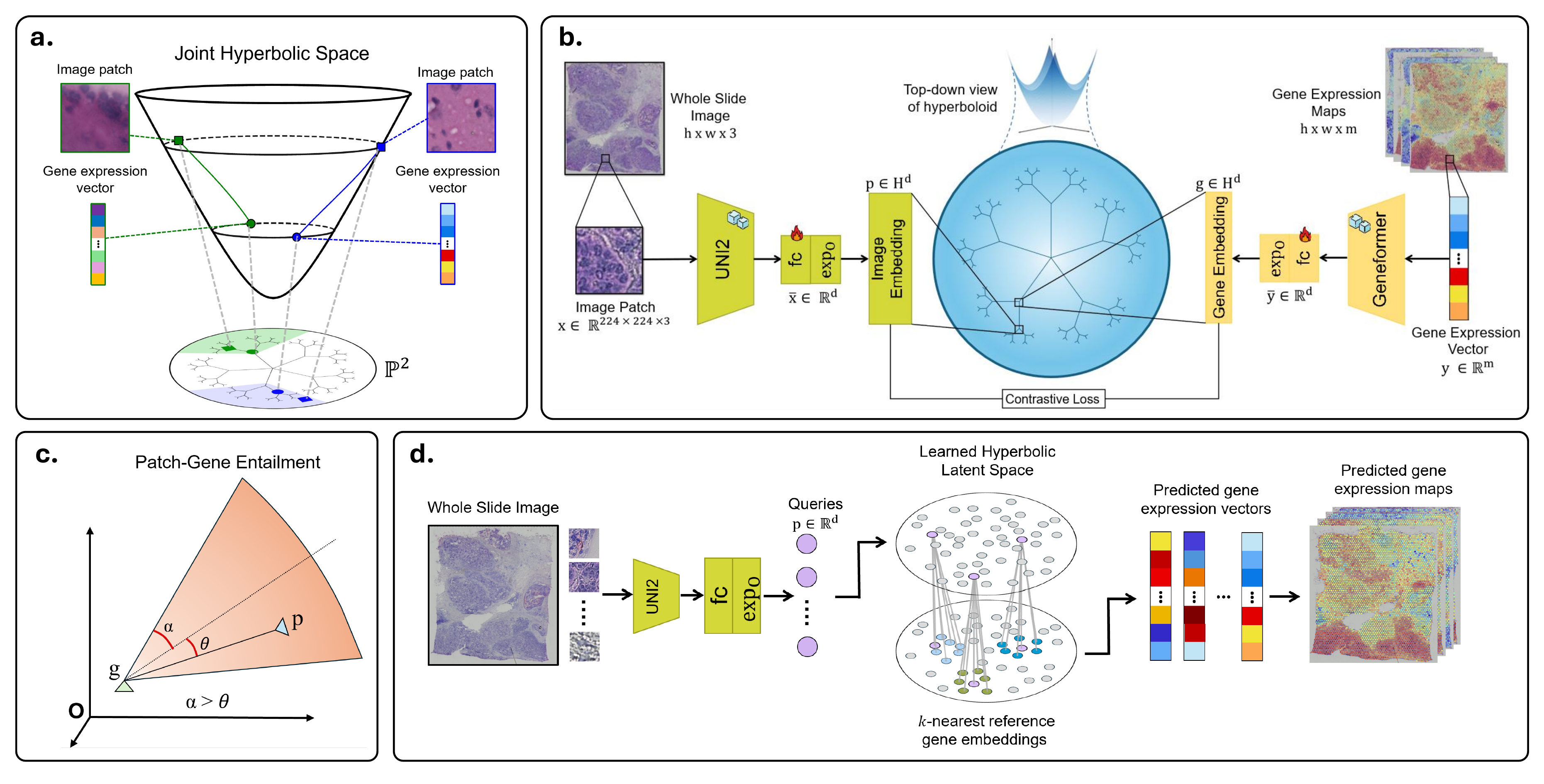}
\caption{(a) \methodname{} learns a joint hierarchical representation of histopathology patches and corresponding gene expression vectors via contrastive learning in hyperbolic space. (b) During training, image patches and paired gene expression vectors are encoded with \uniname{} and Geneformer and projected into a shared hyperbolic space. (c) A patch–gene entailment objective models the asymmetric relationship between modalities. (d) In inference, a patch is encoded and projected into the learned hyperbolic space, and its prediction is obtained by averaging the $k$ nearest gene embeddings.} 
\label{fig:overview}
\end{figure*}

\section{Introduction}
\label{sec:intro}
Spatial Transcriptomics (ST) enables the measurement of gene expression within the spatial architecture of tissues \cite{marx2021method}, effectively aligning histological appearance with local gene expression profiles and providing a rich view of tissue organization and disease mechanisms \cite{robles2023spatial,liu2024spatial}. However, its high cost, specialized equipment requirements, and susceptibility to technical noise and missing measurements limit its scalability and the development of robust Artificial Intelligence (AI) models \cite{jin2024advances,wan2023integrating,li2022sd2}.

Numerous studies have explored computer vision methods to predict spatially resolved gene expression from histopathology images \cite{STNet,Hist2ST,song2024predicting}. Existing approaches can be broadly categorized into architecture-focused and representation-focused methods. Architecture-focused models, such as ST-Net \cite{STNet}, HisToGene \cite{HistToGene}, EGN \cite{EGN}, EGGN \cite{EGGN}, and MERGE \cite{MERGE}, refine network designs to capture local morphology and long-range dependencies, while BLEEP \cite{BLEEP} introduces multimodal contrastive alignment in a shared Euclidean space.
More recently, NH²2ST \cite{NH2ST} employs a dual-branch architecture combining cross-attention and contrastive learning: a query branch for paired histology patches and gene expression vectors, and a neighbor branch to capture multi-scale spatial context.


In these methods, performance improvements have largely plateaued, suggesting that further progress may depend more on representation geometry. Consequently, representation-focused methods such as TRIPLEX \cite{TRIPLEX} and M2TGLGO \cite{M2TGLGO} incorporate multi-resolution and biological functional priors, and HAGE \cite{HAGE} further injects gene co-expression relationships as prior knowledge into the image encoder via cross-attention, aligning gene-guided image and expression embeddings through contrastive learning, yet all still rely on Euclidean embeddings.

Biological systems are inherently hierarchical: gene expression constrains tissue organization through structured regulatory pathways that give rise to morphology \cite{hypothesis}. However, most existing approaches fail to model this hierarchy explicitly. Although DELST \cite{DELST2025} introduces hyperbolic entailment, its hyperbolic component mainly serves as a regularizer, with multimodal alignment still performed in Euclidean space. To overcome these limitations, we propose Hyperbolic Contrastive Learning with Entailment for Spatial Transcriptomics (\methodname{}), a hierarchically structured contrastive framework that predicts gene expression from histopathology images using hyperbolic representations.

As shown in Fig. \ref{fig:overview}a, HyCLoST projects embeddings into a joint hyperbolic space, which naturally captures hierarchical relationships and enhances contrastive alignment. Inspired by \cite{MERU2024},  HyCLoST also incorporates an entailment loss to enforce a directional relationship between gene expression and image embeddings, reflecting the biological asymmetry in which molecular profiles give rise to tissue morphology. We build upon pretrained foundation models, \uniname{} \cite{UNI2024} and Geneformer \cite{Geneformer2023}, to extract high-quality unimodal features. Furthermore, we address inconsistencies in prior evaluations by re-evaluating recent methods within the unified SpaRED benchmark \cite{spared}, which standardizes comparison across 26 public datasets.

To summarize, our main contributions are: (1) We introduce HyCLoST, a hyperbolic contrastive learning framework that leverages hierarchical geometry to improve multimodal alignment and prediction accuracy in ST. (2) We propose a gene-to-image entailment loss that captures the directional relationships between gene expression profiles and tissue morphology. (3) We unify gene expression prediction evaluation under a standardized benchmark. We evaluate our model on the 26 datasets of the SpaRED benchmark \cite{spared}, achieving an average 6\% reduction in Mean Squared Error (MSE) and an 8\% increase in Pearson Correlation Coefficient (PCC) compared to prior methods. To promote further research on ST, our source code is publicly available at  \url{https://github.com/dvegaa00/HyCLoST}.

\section{Method}

The general workflow of \methodname{} is illustrated in Fig. \ref{fig:overview}b. We encode histopathology patches and gene expression profiles using pretrained foundation models (\uniname{} and Geneformer), project them to a shared latent space, and map them to a joint hyperbolic representation. Training optimizes two objectives: a symmetric contrastive loss using negative hyperbolic distance as logits, and an entailment loss that enforces directional gene-to-image consistency via semantic cones.


As shown in Fig.~\ref{fig:overview}d, during inference, \methodname{} encodes test patches and projects them into the learned hyperbolic space. We obtain predictions via retrieval: for each test patch embedding, the model identifies the top-$k$ nearest training gene embeddings under the Lorentzian inner product and averages their corresponding gene expression vectors. This non-parametric strategy is well suited to SpaRED's heterogeneity across tissues and species, as retrieval adapts locally to nearby learned gene-expression contexts instead of forcing a single parametric decoder to compress diverse output distributions. At scale, embeddings are precomputed once per dataset, so KNN cost grows linearly with dataset size and remains lightweight relative to feature encoding.

\subsection{Data encoding and hyperbolic mapping}

\methodname{} uses the Lorentz model \cite{nickel2018learning} to represent hyperbolic geometry, where an $n$-dimensional hyperbolic space is embedded in the upper sheet of a two-sheeted hyperboloid in $\mathbb{R}^{n+1}$. Each vector $u \in \mathbb{R}^{n+1}$ is represented as $[u_{\text{space}},\, u_{\text{time}}]$, with $u_{\text{space}} \in \mathbb{R}^{n}$ and $u_{\text{time}} \in \mathbb{R}$ denoting the spatial and time-like components \cite{ratcliffe2006foundations}.The Lorentzian inner product between $u, v \in \mathbb{R}^{n+1}$ is defined as:

\begin{equation}
    \label{eq:inner_prod}
    \left\langle u,v \right\rangle_{\mathbb{L}}=-u_{time}v_{time}+\sum_{j=1}^{d}{u_{j}v_{j}},
\end{equation}

where $\sum_{j=1}^{d}{u_{j}v_{j}}$, represents the Euclidean inner product $\left\langle u,v \right\rangle$ of $u_{\text{space}}$ and $v_{\text{space}}$. Moreover, the hyperboloid curvature $-c$, is defined as:

\begin{equation}
    \label{eq:eq_two}
    \mathbb{L}^n = \left\{\, u \in \mathbb{R}^{n+1} \;:\; \langle u, u \rangle_{\mathbb{L}} = -\frac{1}{c},   u_{\text{time}} = \sqrt{\frac{1}{c} + \|u_{\text{space}}\|^{2}}. \,\right\}, \qquad c > 0,
\end{equation}

where $\| \cdot \|$ is the Euclidean norm. The curvature $c$ is learned via $c=\exp(\tilde{c})$, 
with $\tilde{c}$ optimized jointly with the network. Thus, given a Euclidean embedding $z \in \mathbb{R}^n$, we map it from the tangent space at the origin to the hyperboloid via the exponential map (Eq.~\ref{eq:expo_map}), with the time component determined by the hyperboloid constraint in Eq.~\ref{eq:eq_two}:
\begin{equation}
    \label{eq:expo_map}
    z_{\text{space}} = \exp_{\mathbf{O}}(z) =
    \frac{\sinh\!\left(\sqrt{c}\,\|z\|\right)}{\sqrt{c}\,\|z\|}\, z
\end{equation}
To construct the patch-gene hyperbolic space, each histopathology patch  $x_i \in \mathbb{R}^{224 \times 224 \times 3}$ and its corresponding gene expression vector $y_i \in \mathbb{R}^{m}$ are encoded using pretrained models \uniname{} $f(\cdot)$ and Geneformer $h(\cdot)$, yielding embeddings $\hat{x}_i \in \mathbb{R}^{1536}$ and $\hat{y}_i \in \mathbb{R}^{768}$. These representations are linearly projected to a shared latent space of dimension $d=512$, producing $\bar{x}_i, \bar{y}_i \in \mathbb{R}^{512}$. The projected embeddings are then mapped via Eq.~\ref{eq:expo_map} to the hyperbolic space, resulting in hyperbolic representations $p_i, g_i \in H^{512}$.


\subsection{Contrastive learning with hyperbolic representations}
After projecting image and gene embeddings into $\mathrm{H}^d$, \methodname{} aligns both modalities using a symmetric contrastive objective based on hyperbolic geodesic distance. Following \cite{MERU2024}, the distance between embeddings $p_i$ and $g_i$ is computed as:
\begin{equation}
    d_{\mathbb{L}}(p_{i},g_{i})=\frac{1}{\sqrt{c}}\cosh^{-1}(-c \left\langle p_{i},g_{i} \right\rangle_{\mathbb{L}}).
\end{equation}
We use the negative distance as similarity, defining logits for a given set of $N$ paired samples as:
 $S_{ij}^{(P)}=-d_{\mathbb{L}}(p_{i},g_{j}),  \ S_{ij}^{(G)}=-d_{\mathbb{L}}(g_{i},p_{j})$.

Each patch-gene pair $(p_{i},g_{i})$ constitutes a positive match, while all other pairs in the set act as negatives. Consequently, the symmetric contrastive loss is: 
\begin{equation}
    \mathbb{L}_{con}=\frac{1}{2N}\sum_{i=1}^{N}\left[ \ell(S_{i}^{(P)},i) + \ell(S_{i}^{(G)},i) \right],
\end{equation}

where $\ell(S_{i}^{(P)},i)$ denotes the cross-entropy loss with target $i$ corresponding to the diagonal element. This objective pulls matched embeddings closer while pushing mismatched pairs apart, leveraging hyperbolic geometry to structure hierarchical relationships across modalities.
\subsection{Gene-to-Image entailment}

While the contrastive loss aligns paired image and gene embeddings symmetrically, the relationship between these modalities is inherently asymmetric: the visual appearance of a tissue arises from its underlying gene expression profile \cite{hypothesis}. To model this directional dependency, we introduce an entailment loss inspired by \cite{MERU2024}, which enforces directional containment in hyperbolic space by encouraging image embeddings to lie within the entailment region induced by gene embeddings (gene to image). 
For each patch-gene pair $(p_i,g_i)$ we compute the exterior angle at $g_i$ in the hyperbolic triangle $(\mathbf{O},p_i,g_i)$. 

\begin{equation}
    \text{ext}(g_{i},p_{i})=\cos^{-1}\left( \frac{p_{i,time}+g_{i,time}c\left\langle p_i,g_i \right\rangle_\mathbb{L}}{\left\| g_{i,space} \right\|\sqrt{(c\left\langle p_i,g_i\right\rangle_{\mathbb{L}})^{2}-1}} \right),
\end{equation}

Intuitively, the closer $p_i$ lies to this direction, the more consistent the histological appearance is with its underlying gene expression profile. As shown in Fig.~\ref{fig:overview}c, each gene embedding $g_i$ defines an entailment cone on the hyperboloid, parameterized by a half-aperture angle defined as: 
\begin{equation}
\alpha_i = \sin^{-1}\!\left(\frac{2\, r_{\min}}{\|g_i\|\sqrt{c}} \right),
\end{equation}
where $r_{\min}$ denotes the minimum cone radius (set to 0.1) and $c$ is the hyperbolic curvature. The entailment loss penalizes violations of this geometric constraint:
\begin{equation}
    \mathcal{L}_{ent} =
    \frac{1}{2}
    \sum_{i=1}^{N}
    \max\!\left(0, \theta_i - \alpha_i \right),
\end{equation}
where $\theta_i = \text{ext}(g_{i},p_{i})$.

Minimizing $\mathcal{L}_{ent}$ enforces a containment relation in which gene expression embeddings geometrically ``entail'' their patch counterparts. The overall training objective combines the contrastive and entailment components as
$\mathcal{L} = \mathcal{L}{con} + \mu \mathcal{L}{ent}$,
where $\mu$ controls the relative contribution of the entailment loss during training.

\section{Results}
\label{sec:results}
\subsection{Experimental Framework}
\label{subsec:exp_fr}

\textbf{Datasets}.
\label{subsec:datasets}
We use the 26 publicly available datasets from SpaRED \cite{spared}, comprising 14 human and 12 mouse datasets across nine tissue types. These datasets are standardized for the gene expression prediction task of the top 128 or 32 most spatially autocorrelated genes. Following previous work \cite{spared,lgdist}, we pre-process the datasets using LGDiST \cite{lgdist}. We evaluate gene expression prediction using MSE and PCC computed only on genes directly measured by ST, excluding pre-completed values.

\textbf{State-of-the-Art} 
\label{subsec:sota}
We compare \methodname{} against nine state-of-the-art methods: ST-Net \cite{STNet}, HisToGene \cite{HistToGene}, SEPAL \cite{sepal}, EGN \cite{EGN}, EGGN \cite{EGGN}, BLEEP \cite{BLEEP}, TRIPLEX \cite{TRIPLEX}, MERGE \cite{MERGE}, and M2TGLGO \cite{M2TGLGO}. All methods were implemented following their official repositories.

For M2TGLGO, the original configuration limits training and evaluation to genes with available Gene2Vec embeddings, excluding on average 6 of 128 genes per dataset. We report results for the Gene2Vec-restricted variant (denoted by *) and for a Geneformer-based variant that covers the full gene set (reported without *).  For TRIPLEX, 1-3 genes are predicted as constant across datasets, affecting evaluation. We report these results without *, and additionally provide metrics excluding constant gene predictions (denoted by *).

\textbf{Implementation Details} 
\label{sec:implem_det}
We train on a NVIDIA Quadro RTX 8000 using AdamW \cite{kingma2014adam} (weight decay 0.2, batch size 128). The learning rate follows a linear warm-up (4,000 steps) to $5\times10^{-4}$ with cosine decay over 120,000 total steps. To reduce computation, foundation model embeddings are precomputed and reused across experiments. For inference, we define $k$ as 80.

\begin{table}[h]
\centering
\caption{Average MSE and PCC over the 26 datasets in SpaRED. Symbol * denotes models missing a set of genes due to constraints (Section \ref{subsec:sota}). The best results are highlighted in \textbf{bold}, and the second best are \underline{underlined}. \#Best indicates the number of datasets where each method achieves the best MSE. Avg. Rank denotes the average MSE-based rank across the 26 datasets, as used in the Friedman test. $\checkmark$ denotes statistically significant improvement of \methodname{} over the corresponding baseline in MSE under paired one-sided Wilcoxon tests with Holm–Bonferroni correction ($\alpha=0.05$).}
\setlength{\tabcolsep}{6pt}
\resizebox{0.98\textwidth}{!}{
\begin{tabular}{cccccc}
\hline
\textbf{Model} & \textbf{MSE} $(\downarrow)$ & \textbf{PCC} $(\uparrow)$ & \textbf{\#Best} & \textbf{Avg. Rank} $(\downarrow)$ & \textbf{Sig.} \\ \hline
TRIPLEX \cite{TRIPLEX}        & 2.433 $\pm$ 1.712 & 0.388 $\pm$ 0.208 & 0 & 10.167 & $\checkmark$ \\
HisToGene \cite{HistToGene}      & 1.455 $\pm$ 0.581 & 0.166 $\pm$ 0.126 & 0 & 9.604 & $\checkmark$ \\ 
TRIPLEX* \cite{TRIPLEX}        & 1.126 $\pm$ 0.589 & 0.393 $\pm$ 0.211 & 1 & 7.833 & $\checkmark$ \\ 
MERGE \cite{MERGE}         & 1.123 $\pm$ 0.736 & 0.339 $\pm$ 0.280  & 1 & 6.417 & $\checkmark$ \\
BLEEP \cite{BLEEP}          & 1.106 $\pm$ 0.588 & 0.317 $\pm$ 0.184 & 0 & 8.375 & $\checkmark$ \\
M2TGLGO \cite{M2TGLGO}        & 1.087 $\pm$ 0.417 & 0.174 $\pm$ 0.135 & 0 & 8.208 & $\checkmark$ \\
SEPAL \cite{sepal}          & 0.999 $\pm$ 0.841 & 0.393 $\pm$ 0.181 & \underline{4} & 3.938 & $\times$ \\
EGN \cite{EGN}            & 0.982 $\pm$ 0.558 & 0.365 $\pm$ 0.154 & 2 & 5.021 & $\checkmark$ \\
STNet \cite{STNet}          & 0.974 $\pm$ 0.625 & \underline{0.400} $\pm$ \underline{0.169} & \underline{4} & \underline{3.854} & $\checkmark$ \\
M2TGLGO* \cite{M2TGLGO}        & 0.954 $\pm$ 0.380 & 0.294 $\pm$ 0.193 & 2 & 6.833 & $\checkmark$ \\
EGGN \cite{EGGN}           & \underline{0.953} $\pm$ \underline{0.442} & 0.371 $\pm$ 0.173 & 2 & 4.958 & $\checkmark$ \\ \hline
\cellcolor[HTML]{EFEFEF} \textbf{\methodname{} (Ours)}  
& \cellcolor[HTML]{EFEFEF} \textbf{0.899} $\pm$ \textbf{0.436} 
& \cellcolor[HTML]{EFEFEF} \textbf{0.431} $\pm$ \textbf{0.176} 
& \cellcolor[HTML]{EFEFEF} \textbf{10} 
& \cellcolor[HTML]{EFEFEF} \textbf{2.792} 
& \cellcolor[HTML]{EFEFEF} -- \\ \hline
\end{tabular}}
\label{tab:quanti}
\end{table}

\subsection{Main Results}
\label{subsec:quant_results}

Tab.~\ref{tab:quanti} compares \methodname{} with nine state-of-the-art methods across the 26 SpaRED datasets, resulting in eleven evaluated configurations due to starred variants. \methodname{} achieves the lowest average MSE and highest average PCC, with improvements of 6\% and 8\% over the second-best method, respectively. The improvement on both metrics indicates that \methodname{} accurately predicts absolute expression values and also captures spatially coherent expression patterns aligned with the ground truth. Moreover, statistical analysis confirms significant differences among methods (Friedman test, $\chi^2 = 117.02$, $p < 10^{-19}$). \methodname{} achieved the best average rank (2.79) and significantly outperformed 10 of 11 baseline configurations under paired one-sided Wilcoxon tests with Holm–Bonferroni correction ($\alpha=0.05$).

Although recent approaches report competitive results, our evaluation in SpaRED shows that improvements over simpler models remain modest. In contrast, \methodname{} delivers consistent improvements in both MSE and PCC within an architecturally efficient, geometrically grounded framework. As shown in Tab.~\ref{tab:quanti}, it ranks first in 10 of 26 datasets, demonstrating robust performance across diverse tissues and acquisition settings.

\subsection{Qualitative Results}

\begin{figure*}[t]
\includegraphics[width=1.02\textwidth]{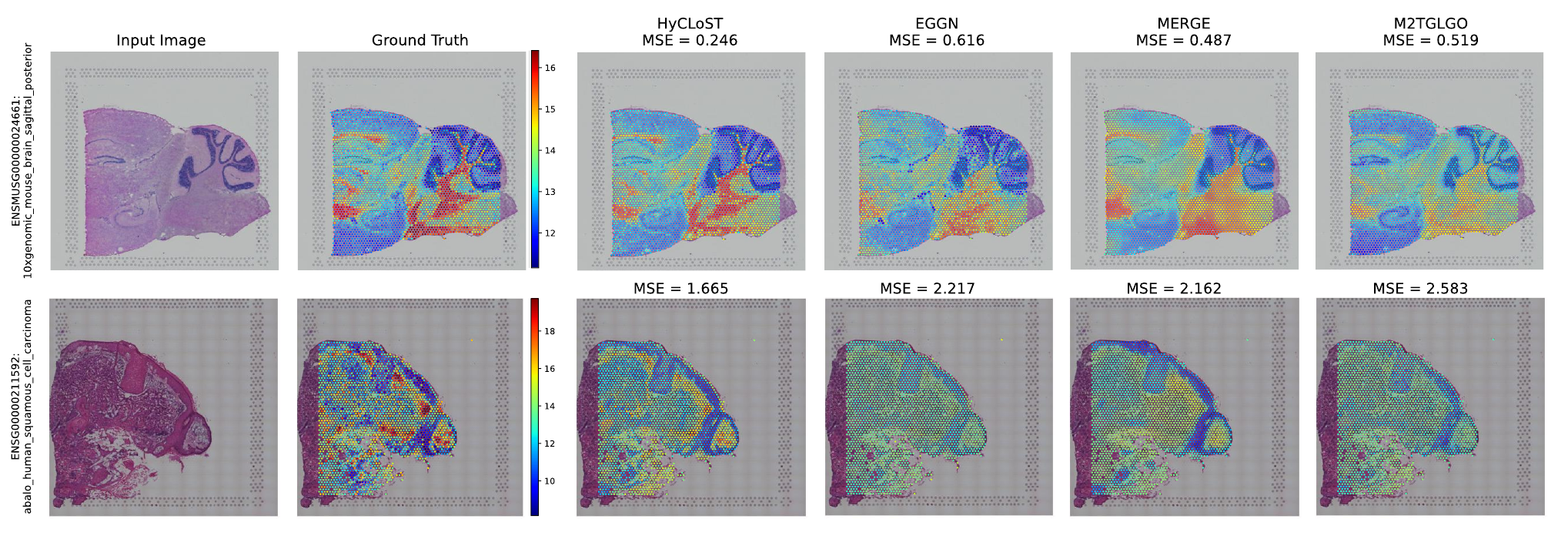}
\caption{Qualitative comparison of spatial gene expression predictions for two genes from two SpaRED datasets. For each gene, we show the input patch, gene expression ground truth, and predictions from \methodname{}, EGGN \cite{EGGN}, MERGE \cite{MERGE}, and M2TGLGO \cite{M2TGLGO}.} 
\label{fig:qualitative}
\end{figure*}

Fig. \ref{fig:qualitative} compares the predictions of \methodname{} with those of the second-best performing model (EGGN, as observed in Sec. \ref{subsec:quant_results}) and the two most recent methods (MERGE and M2TGLGO). The first row shows representative low-error cases. In this example, \methodname{} achieves the lowest MSE and better preserves the dynamic range and spatial patterns of the ground truth, whereas competing models tend to over-smooth or collapse predictions toward uniform values.

In the second row, all methods face more challenging cases and yield higher errors. Nevertheless, \methodname{} consistently achieves the lowest MSE and better captures extreme expression values, particularly in high-expression regions where competing models tend to flatten predictions. Overall, these results highlight \methodname{}’s ability to preserve spatial structure and dynamic range across diverse tissues and genes.

\subsection{Ablation Experiments}
We conduct ablation experiments to validate the geometric and biological principles underlying \methodname{} (Tab.~\ref{tab:ablations}). First, we assess the contribution of the entailment weight loss (Rows 1-3). The results confirm that enforcing a partial-order structure between image and gene embeddings produces a more informative hyperbolic space, thereby improving prediction accuracy.

Next, we evaluate the impact of using foundation models for both the gene and patch encoders. Replacing \uniname{} with a small Vision Transformer (ViT) increases MSE (Row 4 vs. Row 3), indicating less informative patch representations. Next, we keep \uniname{} as the patch encoder and add Geneformer as the gene encoder. This change yields a significant performance improvement (Row 5 vs. Row 3).

Furthermore, we evaluate representation geometry. Replacing the hyperbolic space with a Euclidean one while keeping all other components unchanged substantially increases MSE (Row 6 vs. Row 5). This matched comparison isolates the contribution of the representation space and supports our hypothesis that hyperbolic geometry provides a better inductive bias than Euclidean embeddings for modeling the hierarchical structure induced by gene-regulatory organization.


\begin{table}[] 
\centering 
\caption{Average MSE over the 26 datasets from SpaRED in ablation experiments. The best results are highlighted in \textbf{bold}.} 
\label{tab:ablations} 
\resizebox{0.98\textwidth}{!}{
\begin{tabular}{ccccccc} 
\hline 
\textbf{\#} & \textbf{Entailment Weight} & \textbf{Gene Encoder} & \textbf{Patch Encoder} & \textbf{Entailment Mode} & \textbf{Geometry} & \textbf{MSE} $(\downarrow)$  \\ 
\hline 
1  & 0   & None        & UNI & None  & Hyperbolic & 0.958  \\ 
2  & 0.2 & None        & UNI & Gene  & Hyperbolic & 0.942  \\ 
3  & 0.5 & None        & UNI & Gene  & Hyperbolic & 0.933  \\ 
4  & 0.5   & None        & ViT & Gene  & Hyperbolic & 1.206  \\ 
\textbf{5} & \textbf{0.5} & \textbf{Geneformer} & \textbf{UNI} & \textbf{Gene} & \textbf{Hyperbolic} & \textbf{0.899} \\ 
6  & 0.5 & Geneformer  & UNI & Gene  & Euclidean  & 1.043  \\ 
7  & 0.5 & Geneformer  & UNI & Image & Hyperbolic & 0.919  \\ 
\hline  
\end{tabular}}
\end{table}

Finally, we analyze the directionality of the entailment relation. Comparing the configuration “gene entails image” (Row 5) with “image entails gene” (Row 7) shows that the former achieves a lower MSE. Although the average gap is moderate, the effect is consistent across datasets. Inspecting the per-dataset results underlying Tab. 2 reveals that the gene-to-image configuration attains lower MSE in 21 out of 26 datasets. This supports our analytical hypothesis that gene expression provides a more general molecular description that constrains plausible tissue morphologies, while histological appearance represents an observable outcome within a constrained set of possibilities \cite{hypothesis}.

\section{Conclusions}
In this paper, we present \methodname{}, a novel geometric learning framework for predicting gene expression from histopathology images. \methodname{} builds a joint hyperbolic space to model hierarchical relationships within ST data and guide contrastive alignment. A gene-to-image entailment loss captures the asymmetric dependence of tissue appearance on underlying gene expression, yielding predictions. Experiments on the 26 SpaRED datasets show average improvements of 6\% in MSE and 8\% in PCC over prior methods. Overall, our results indicate that refining the geometry and semantics of the representation space enables more expressive modeling of biological structure, highlighting a promising direction for spatial gene expression prediction.

\begin{credits}
\subsubsection{\ackname} This work was supported by Azure sponsorship credits granted by Microsoft’s AI for Good Research Lab.

\subsubsection{\discintname}
The authors have no competing interests to declare that are
relevant to the content of this article.
\end{credits}

%
%
%
%


\bibliographystyle{splncs04}
\bibliography{Paper-3824}

@article{marx2021method,
  title={Method of the Year: spatially resolved transcriptomics},
  author={Marx, Vivien},
  journal={Nature methods},
  volume={18},
  number={1},
  pages={9--14},
  year={2021},
  publisher={Nature Publishing Group US New York}
}

@article{robles2023spatial,
  title={Spatial transcriptomics: emerging technologies in tissue gene expression profiling},
  author={Robles-Remacho, Agust{\'\i}n and Sanchez-Martin, Rosario M and Diaz-Mochon, Juan J},
  journal={Analytical Chemistry},
  volume={95},
  number={42},
  pages={15450--15460},
  year={2023},
  publisher={ACS Publications}
}

@article{jin2024advances,
  title={Advances in spatial transcriptomics and its applications in cancer research},
  author={Jin, Yang and Zuo, Yuanli and Li, Gang and Liu, Wenrong and Pan, Yitong and Fan, Ting and Fu, Xin and Yao, Xiaojun and Peng, Yong},
  journal={Molecular Cancer},
  volume={23},
  number={1},
  pages={129},
  year={2024},
  publisher={Springer}
}

@article{wan2023integrating,
  title={Integrating spatial and single-cell transcriptomics data using deep generative models with SpatialScope},
  author={Wan, Xiaomeng and Xiao, Jiashun and Tam, Sindy Sing Ting and Cai, Mingxuan and Sugimura, Ryohichi and Wang, Yang and Wan, Xiang and Lin, Zhixiang and Wu, Angela Ruohao and Yang, Can},
  journal={Nature Communications},
  volume={14},
  number={1},
  pages={7848},
  year={2023},
  publisher={Nature Publishing Group UK London}
}

@article{li2022sd2,
  title={SD2: spatially resolved transcriptomics deconvolution through integration of dropout and spatial information},
  author={Li, Haoyang and Li, Hanmin and Zhou, Juexiao and Gao, Xin},
  journal={Bioinformatics},
  volume={38},
  number={21},
  pages={4878--4884},
  year={2022},
  publisher={Oxford University Press}
}

@article{song2024predicting,
  title={Predicting spatially resolved gene expression via tissue morphology using adaptive spatial GNNs},
  author={Song, Tianci and Cosatto, Eric and Wang, Gaoyuan and Kuang, Rui and Gerstein, Mark and Min, Martin Renqiang and Warrell, Jonathan},
  journal={Bioinformatics},
  volume={40},
  number={Supplement\_2},
  pages={ii111--ii119},
  year={2024},
  publisher={Oxford University Press}
}

@misc{MERU2024,
      title={Hyperbolic Image-Text Representations}, 
      author={Karan Desai and Maximilian Nickel and Tanmay Rajpurohit and Justin Johnson and Ramakrishna Vedantam},
      year={2024},
      eprint={2304.09172},
      archivePrefix={arXiv},
      primaryClass={cs.CV},
      url={https://arxiv.org/abs/2304.09172}, 
}

@misc{DELST2025,
      title={DELST: Dual Entailment Learning for Hyperbolic Image-Gene Pretraining in Spatial Transcriptomics}, 
      author={Xulin Chen and Junzhou Huang},
      year={2025},
      eprint={2503.00804},
      archivePrefix={arXiv},
      primaryClass={cs.CV},
      url={https://arxiv.org/abs/2503.00804}, 
}

@article{Geneformer2023, 
        title={Transfer learning enables predictions in Network Biology}, volume={618}, 
        DOI={10.1038/s41586-023-06139-9}, 
        number={7965}, 
        journal={Nature}, 
        author={Theodoris, Christina V. and Xiao, Ling and Chopra, Anant and Chaffin, Mark D. and Al Sayed, Zeina R. and Hill, Matthew C. and Mantineo, Helene and Brydon, Elizabeth M. and Zeng, Zexian and Liu, X. Shirley and et al.}, 
        year={2023}, 
        month={May}, 
        pages={616–624}}

@article{UNI2024,
  title={Towards a General-Purpose Foundation Model for Computational Pathology},
  author={Chen, Richard J and Ding, Tong and Lu, Ming Y and Williamson, Drew FK and Jaume, Guillaume and Chen, Bowen and Zhang, Andrew and Shao, Daniel and Song, Andrew H and Shaban, Muhammad and others},
  journal={Nature Medicine},
  publisher={Nature Publishing Group},
  year={2024}
}

@inproceedings{lgdist,
  title={Latent Gene Diffusion for Spatial Transcriptomics Completion},
  author={C{\'a}rdenas, Paula and Manrique, Leonardo and Vega, Daniela and Ruiz, Daniela and Arbel{\'a}ez, Pablo},
  booktitle={Proceedings of the IEEE/CVF International Conference on Computer Vision},
  pages={901--910},
  year={2025}
}

@article{STNet,
   doi = {10.1038/s41551-020-0578-x},
   issn = {2157846X},
   issue = {8},
   journal = {Nature Biomedical Engineering},
   title = {Integrating spatial gene expression and breast tumour morphology via deep learning},
   author  = {He, Binbin and Bergenstr{\aa}hle, Ludvig and Stenbeck, Linus and Abid, Areeba and Andersson, Alma and Borg, {\AA}ke and Maaskola, Jonas and Lundeberg, Joakim and Zou, James},
   volume = {4},
   year = {2020}
}

@article{HistToGene,
   author = {Minxing Pang and Kenong Su and Mingyao Li},
   journal = {bioRxiv},
   title = {Leveraging information in spatial transcriptomics to predict super-resolution gene expression from histology images in tumors},
   year = {2021}
}

@article{Hist2ST,
   author = {Yuansong Zeng and Zhuoyi Wei and Weijiang Yu and Rui Yin and Yuchen Yuan and Bingling Li and Zhonghui Tang and Yutong Lu and Yuedong Yang},
   doi = {10.1093/bib/bbac297},
   issn = {14774054},
   issue = {5},
   journal = {Briefings in Bioinformatics},
   title = {Spatial transcriptomics prediction from histology jointly through Transformer and graph neural networks},
   volume = {23},
   year = {2022}
}

@inproceedings{EGN,
   author = {Yan Yang and Md Zakir Hossain and Eric A. Stone and Shafin Rahman},
   doi = {10.1109/WACV56688.2023.00501},
   booktitle = {Proceedings - 2023 IEEE Winter Conference on Applications of Computer Vision, WACV 2023},
   title = {Exemplar Guided Deep Neural Network for Spatial Transcriptomics Analysis of Gene Expression Prediction},
   year = {2023}
}

@ARTICLE{EGGN,
        title = "{Spatial transcriptomics analysis of gene expression prediction using exemplar guided graph neural network}",
       author = {Yan Yang and Md Zakir Hossain and Eric Stone and Shafin Rahman},
      journal = {Pattern Recognition},
         year = 2024,
        month = jan,
       volume = {145},
          eid = {109966},
        pages = {109966},
          doi = {10.1016/j.patcog.2023.109966},
       adsurl = {https://ui.adsabs.harvard.edu/abs/2024PatRe.14509966Y}
}

@misc{BLEEP,
      title={Spatially Resolved Gene Expression Prediction from H\&E Histology Images via Bi-modal Contrastive Learning}, 
      author={Ronald Xie and Kuan Pang and Sai W. Chung and Catia T. Perciani and Sonya A. MacParland and Bo Wang and Gary D. Bader},
      year={2023},
      eprint={2306.01859},
      archivePrefix={arXiv},
      primaryClass={cs.CV},
      url={https://arxiv.org/abs/2306.01859}, 
}

@INPROCEEDINGS{M2TGLGO,
  author={Shi, Hang and Chi, Changxi and Wan, Peng and Zhang, Daoqiang and Shao, Wei},
  booktitle={2025 IEEE/CVF Conference on Computer Vision and Pattern Recognition (CVPR)}, 
  title={Multi-modal Topology-embedded Graph Learning for Spatially Resolved Genes Prediction from Pathology Images with Prior Gene Similarity Information}, 
  year={2025},
  volume={},
  number={},
  pages={20810-20819},
  doi={10.1109/CVPR52734.2025.01938}}

@misc{MERGE,
      title={MERGE: Multi-faceted Hierarchical Graph-based GNN for Gene Expression Prediction from Whole Slide Histopathology Images}, 
      author={Aniruddha Ganguly and Debolina Chatterjee and Wentao Huang and Jie Zhang and Alisa Yurovsky and Travis Steele Johnson and Chao Chen},
      year={2025},
      eprint={2412.02601},
      archivePrefix={arXiv},
      primaryClass={cs.CV},
      url={https://arxiv.org/abs/2412.02601}, 
}

@misc{TRIPLEX,
      title={Accurate Spatial Gene Expression Prediction by integrating Multi-resolution features}, 
      author={Youngmin Chung and Ji Hun Ha and Kyeong Chan Im and Joo Sang Lee},
      year={2024},
      eprint={2403.07592},
      archivePrefix={arXiv},
      primaryClass={cs.CV},
      url={https://arxiv.org/abs/2403.07592}, 
}

@article{kingma2014adam,
  title={Adam: A method for stochastic optimization},
  author={Kingma, Diederik P and Ba, Jimmy},
  journal={arXiv preprint arXiv:1412.6980},
  year={2014}
}

@inproceedings{sepal,
  title={SEPAL: spatial gene expression prediction from local graphs},
  author={Mejia, Gabriel and C{\'a}rdenas, Paula and Ruiz, Daniela and Castillo, Angela and Arbel{\'a}ez, Pablo},
  booktitle={Proceedings of the IEEE/CVF International Conference on computer vision},
  pages={2294--2303},
  year={2023}
}

@inproceedings{spared,
  title={Enhancing Gene Expression Prediction from Histology Images with Spatial Transcriptomics Completion},
  author={Mejia, Gabriel and Ruiz, Daniela and C{\'a}rdenas, Paula and Manrique, Leonardo and Vega, Daniela and Arbel{\'a}ez, Pablo},
  booktitle={International Conference on Medical Image Computing and Computer-Assisted Intervention},
  pages={91--101},
  year={2024},
  organization={Springer}
}

@inproceedings{nickel2018learning,
  title={Learning continuous hierarchies in the lorentz model of hyperbolic geometry},
  author={Nickel, Maximillian and Kiela, Douwe},
  booktitle={International conference on machine learning},
  pages={3779--3788},
  year={2018},
  organization={PMLR}
}

@book{ratcliffe2006foundations,
  title={Foundations of hyperbolic manifolds},
  author={Ratcliffe, John G},
  year={2006},
  publisher={Springer}
}

@article{liu2024spatial,
  title={Spatial multi-omics: deciphering technological landscape of integration of multi-omics and its applications},
  author={Liu, Xiaojie and Peng, Ting and Xu, Miaochun and Lin, Shitong and Hu, Bai and Chu, Tian and Liu, Binghan and Xu, Yashi and Ding, Wencheng and Li, Li and others},
  journal={Journal of Hematology and Oncology},
  volume={17},
  number={1},
  pages={72},
  year={2024},
  publisher={Springer}
}

@article{hypothesis,
  title={Histology image analysis of 13 healthy tissues reveals molecular-histological correlations},
  author={Gao, Yi and Liang, Jianwen and Tian, Mu and Deng, Wenjiang and Wang, Lei and Tao, Siyuan and Mou, Tian},
  journal={Scientific Reports},
  volume={15},
  number={1},
  pages={26812},
  year={2025},
  publisher={Nature Publishing Group UK London}
}

@inproceedings{NH2ST,
  title={Spatially Gene Expression Prediction using Dual-Scale Contrastive Learning},
  author={Qu, Mingcheng and Wu, Yuncong and Di, Donglin and Gao, Yue and Su, Tonghua and Song, Yang and Fan, Lei},
  booktitle={International Conference on Medical Image Computing and Computer-Assisted Intervention},
  year={2025},
  publisher={Springer}
}

@inproceedings{HAGE,
  title={{HAGE}: Hierarchical Alignment Gene-Enhanced Pathology Representation Learning with Spatial Transcriptomics},
  author={Dang, Thao M. and Li, Haiqing and Guo, Yuzhi and Ma, Hehuan and Jiang, Feng and Miao, Yuwei and Zhou, Qifeng and Gao, Jean and Huang, Junzhou},
  booktitle={International Conference on Medical Image Computing and Computer-Assisted Intervention},
  year={2025},
  publisher={Springer}
}

\end{document}